\documentclass[mlabstract,twocolumn]{jmlr}
\usepackage{longtable}% for long tables
\usepackage{listings}
\usepackage{booktabs}
\usepackage{multirow}
\usepackage{subfiles}
\usepackage{booktabs}
\usepackage[load-configurations=version-1]{siunitx} % newer version

\theorembodyfont{\upshape}
\theoremheaderfont{\scshape}
\theorempostheader{:}
\theoremsep{\newline}

\jmlrvolume{}
\firstpageno{1}
\jmlryear{2022}
\jmlrworkshop{Machine Learning for Health (ML4H) 2022}

\title{HealthE: Classifying  Entities in Online Textual Health Advice}

  \author{\Name{Joseph Gatto}\thanks{Equal Contribution} \Email{joseph.m.gatto.gr@dartmouth.edu} \\
  \Name{Parker Seegmiller}\footnotemark[1] \Email{matthew.p.seegmiller.gr@dartmouth.edu } \\
  \Name{Garrett Johnston} \Email{garrett.m.johnston.22@dartmouth.edu }\\
  \Name{Sarah M. Preum} \Email{sarah.masud.preum@dartmouth.edu }\\}

\begin{document}

\maketitle

% \input{0abstract.tex}

% \begin{keywords}
% List of keywords
% \end{keywords}

% \subfile{Sections/0abstract}
\begin{abstract}
The processing of entities in natural language is essential to many medical NLP systems. Unfortunately, existing datasets vastly under-represent the entities required to model public health relevant texts such as \textit{health advice} often found on sites like WebMD. People rely on such information for personal health management and clinically relevant decision making. In this work, we release a new annotated dataset, \textbf{HealthE}, consisting of 6,756 health advice. HealthE has a more granular label space compared to existing medical NER corpora and contains annotation for diverse health phrases. 

Additionally, we introduce a new health entity classification model, \textit{EP S-BERT}, which leverages textual context patterns in the classification of entity classes. \textit{EP S-BERT} provides a 4-point increase in F1 score over the nearest baseline and a 34-point increase in F1 when compared to off-the-shelf medical NER tools trained to extract disease and medication mentions from clinical texts. All code and data are publicly available on Github.

\end{abstract}

\section{Introduction}
\begin{figure*}[htbp]
    \centering
    \includegraphics[scale=0.75]{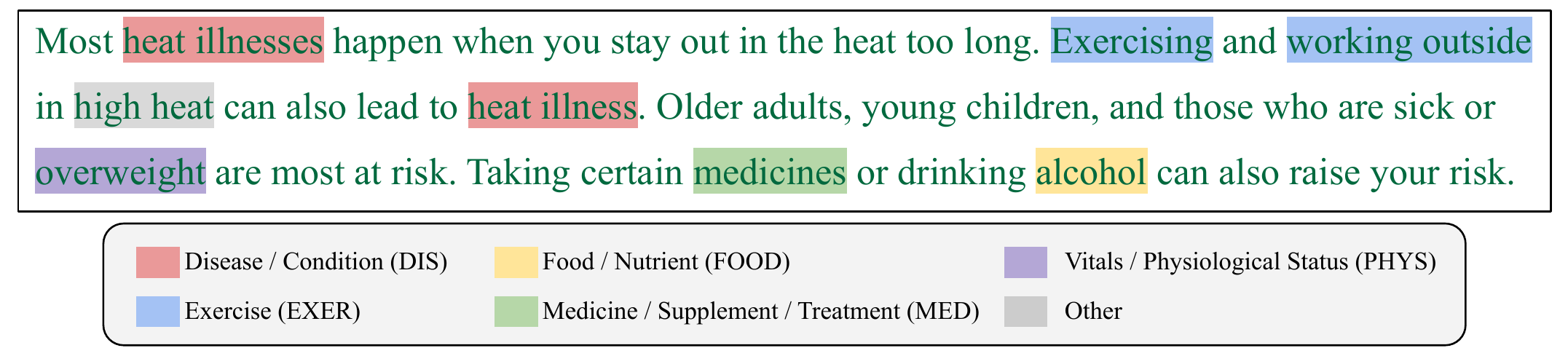}
    \caption{Sample from the HealthE dataset with entities labeled from each class in the HealthE label space.  }
    \label{data_sample}
\end{figure*}

\textbf{Background and knowledge gap}: Health information or \textit{health advice} found on patient education sites (e.g., WebMD, Mayo Clinic, CDC website, American Diabetes Association) plays an important role in improving health literacy, medical information search, and patient empowerment \cite{calixte2020social, masoni2013pharmacovigilance, kubb2020online}. Entity classification from such online textual health advice can be essential to various downstream medical natural language processing (medNLP) tasks including but not limited to misinformation detection \cite{swire2019public}, medical dialogue systems \cite{chintagunta2021medically}, and patient-centric information tools \cite{dai2020ginger, preclude, beaunoyer2017understanding}.

% medical knowledge graph construction \cite{LI2020101817}
% adverse drug event detection \cite{pmlr-v90-chapman18a}. 

% The classification of textual entities plays an important role in various medical NLP tasks including medical knowledge graph construction \cite{LI2020101817}, medical dialogue systems \cite{chintagunta2021medically}, and adverse drug event detection \cite{pmlr-v90-chapman18a}. \textcolor{purple}{These works are crucial to tools which aim to aggregate public health information and promote health literacy. }

Unfortunately, off-the-shelf entity classifiers cover only a small subset of the entity classes commonly found in online textual health advice, e.g., disease, drug, side effects. They often overlook entities that are critical to computationally represent health advice, such as food, exercise, and  physiological status unrelated to side effects. This is because existing works largely focus on the extraction of entities found in either technical content  (i.e., electronic health record (EHR) notes and biomedical literature \cite{Neumann2019ScispaCyFA, biobert}) or lay-person generated content (i.e., social media \cite{socialmedia_ner}). 
% Health information or \textit{health advice} found on  patient education sites (e.g., WebMD, Mayo Clinic, CDC website, American Diabetes Association) are vastly under-represented in medical entity extraction solutions. 
% Health texts such as \textit{health advice} often found on WebMD are vastly underrepresented in medical entity extraction pipelines. 
% Understanding the full scope of terms in online health advice, with such entity classes including food, exercise, physiological status, etc., are crucial to the modeling of intermediate level health texts such as health advice. 
% This is important towards the development of patient-centric AI tools which can both assess and compare online health information \cite{dai2020ginger, preclude} [][]. 

% \textcolor{red}{need a line about why this is important: e.g., develop patient-centric, disease specific information assistant/tools, compare and assess online health information [cite papers on medical claims and misinformation].}

\textbf{Our solution:} To address this knowledge gap, we make the following contributions in this paper. We introduce a new annotated dataset and model for medical entity classification from public health related text targeting laypersons. The dataset, \textbf{HealthE}, contains 6,756 human-labeled health advice texts from numerous websites including \textit{WebMD}, \textit{MedlinePlus}, and several disease specific popular health websites. 
% \textcolor{purple}{HealthE is annotated for both health entity extraction} \textcolor{blue}{Advice texts in HealthE are labeled for health entity classification}. 
We explore the problem of health entity classification in this paper to classify  entities into one of six classes: (i) Food / Nutrient, (ii) Disease / Condition, (iii) Medicine / Supplement / Treatment, (iv) Exercise, (v) Vitals / Physiological Status, and (vi) Other. The annotation format of HealthE is also compatible for medical named entity recognition (NER) task. Unlike existing medical NER tools, however, this class set is representative of public health related texts providing a more granular level of text annotation. 
% \textcolor{red}{Non-technical texts such as \textit{health advice} are vastly underrepresented in medical entity extraction pipelines. ([PS] here we can maybe list examples of where the health advice text is coming from - web sources. Everyone knows webMD, for example. These are underutilized resources!)} Understanding the full scope of terms in health texts, with such entity classes including food, exercise, physiological status, etc., are crucial to the modeling of intermediate level health texts such as medical advice statements. 

% \textcolor{red}{([PS] I think we should highlight here that our dataset can be used for NER and ESE in addition to medical entity classification. We choose to approach the med entity classification task with EP S-BERT but we provide baselines for the other tasks (at least NER)} 

We also introduce Entity/Pattern S-BERT (\textit{EP S-BERT}) for medical entity classification. \textit{EP S-BERT} is a transformer-based architecture which leverages textual context patterns, as is commonly utilized in Entity Set Expansion \cite{ESE1, ESE2}, to predict health entity classes. EP S-BERT provides a 34\% increase in F1 score over the nearest Medical NER baseline when used off-the-shelf, and a 4\% increase in performance over the nearest baseline fine-tuned to HealthE. 

% \textcolor{purple}{Our contributions are as follows. (i) We introduce \textbf{HealthE}, a new dataset for health entity classification on intermediate level health advice texts. (ii) We propose \textit{EP S-BERT}, which effectively leverages context patterns to predict a class of relevant health entities. }

% \textcolor{red}{([PS] I would also say challenging for HealthE, since that's our main contribution/task)} and identifying future work.

% \begin{itemize}
%     \item 
%     \item 
%     % \item We perform an in-depth error analysis of EP S-BERT and identify two core challenges and future work on this task. 
% \end{itemize}

% \subfile{2related_work}
\section{Methods}

\subsection{Data Collection and Annotation}
\label{dataset_sec}
HealthE is a dataset of 6,756 textual health advice statements. A sample with annotation from all classes is highlighted in Figure \ref{data_sample}. For each sample, objects or phrases were annotated as one of the displayed 6 classes. For example, in Figure \ref{data_sample}, the entity ``overweight" is labeled Physiological Status, which refers to symptoms, behaviors, organs, and general state of health/being. Other class names are generally self-explanatory, with full annotation details available in the appendix.  

Samples were collected from various online sources of information pertaining to health and general well-being including WebMD, Medline Plus, CDC, COVID Protocols, Yahoo! Health, MayoClinic, as well as 8 mobile health applications. For a detailed view of the distribution of health advice topics from which entities were extracted, please refer to Section \ref{apd:second} in the appendix.

\begin{figure*}
    \centering
    \includegraphics[width=\textwidth]{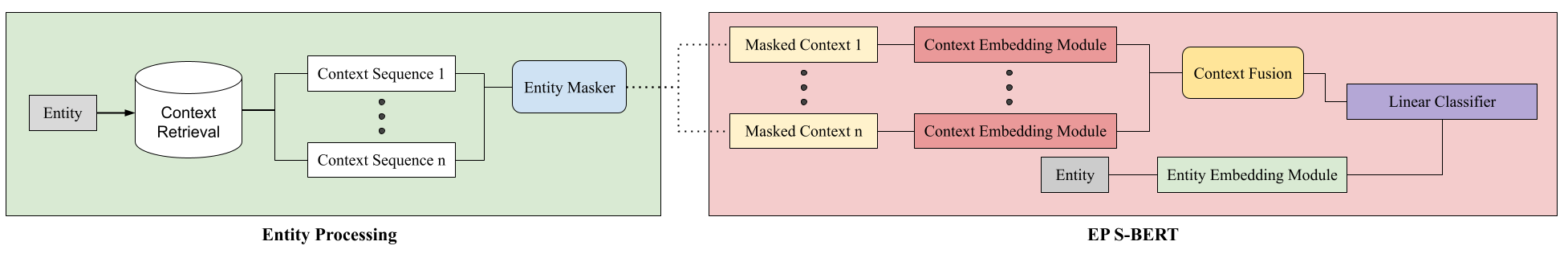}
    \caption{End-to-End health entity classification pipeline for our model EP S-BERT. }
    \label{fig:architecture}
\end{figure*}

\subsection{Entity/Pattern S-BERT}

\textit{EP S-BERT} is based on the Sentence-BERT (S-BERT) architecture \cite{reimers-2019-sentence-bert} which outputs a single vector representation for a given textual input. 
% Specifically, given some set of tokens $X = \{t_1, t_2, \dots, t_n\}$, an S-BERT model, $S(X)$ returns an embedding $\textbf{X}^{\prime} \in R^{d}$ where $d$ is the dimension of the textual embedding. 
The choice to use S-BERT, a sequence encoder, as opposed to a word embedding model is motivated by the large quantity of multi-token entities (e.g. ``exposure to nature", ``drinking enough fluids", "ringing in the ears") found in HealthE. We additionally use S-BERT to encode the context patterns of a given entity. Encoding entity context helps classify multi-meaning context-dependent entities. For example, if we wish to classify the entity “liver” without any context, it is not apparent whether it refers to animal liver that someone may eat, or the organ. However, if we observe that “liver” occurs within patterns such as “beef \_ or chicken”, “chicken \_ will help”, “kidneys , \_ , dairy”, it becomes apparent that we should classify “liver” as FOOD. To generate the context embedding for an entity $E$, we consider the set of advice texts $C = \{h_1, h_2, \dots, h_n\}$ where $h_i$ is an advice text in HealthE containing $E$. $C$ is processed by replacing all entity tokens $E$ with a mask token $[MASK]$, leaving only the entities' context. Formally, this transforms a given health advice statement into $h_i = \{t_1, t_2, \dots, [MASK] , \dots, t_n\}$, retaining all tokens $t_i$ except those pertaining to the entity $E$.   Finally, we embed each text in $C$ using S-BERT and use the mean of all context embeddings as the entity pattern embedding. The concatenation of the entity and pattern embeddings are then fed into a linear classifier. A visualization of this pipeline is shown in Figure \ref{fig:architecture}.

\begin{table}[!h]
\centering 
\begin{tabular}{lc}
\textbf{Class} & \textbf{Number of Samples} \\
\toprule
MED & 2011 \\
DIS & 1162 \\
FOOD & 905 \\
PHYS & 871 \\
EXER & 215 \\
OTH & 254
\end{tabular}
\caption{HealthE Data Distribution}
\end{table}

\section{Experiments and Results}
\begin{table*}[t]
  \centering
    \begin{tabular}{lcccccc|c}
      Model            & DIS  & MED  & FOOD & EXER & PHYS & OTH & W/AVG\\
    \hline
    GloVE              & 0.78 & 0.85 & 0.88 & 0.77 & 0.71 & 0.40 & 0.79 \\
    BERT               & 0.77 & 0.83 & 0.83 & 0.78 & 0.70 & 0.41 & 0.77 \\
    Bio+Clinical BERT  & 0.80 & 0.84 & 0.79 & 0.71 & 0.71 & 0.39 & 0.77 \\
    S-BERT             & 0.79 & 0.86 & 0.87 & 0.84 & 0.71 & 0.36 & 0.80 \\
    \hline 
    SciSpacy           & 0.59 & 0.44 & -    & -    & -    & -   &  0.51    \\
    BioBERT NER        & 0.28 & 0.29 & -    & -    & -    & -   &  0.28    \\
    \hline 
    EP S-BERT          & 0.81 & 0.89 & \textbf{0.92} & \textbf{0.88} & 0.75 & 0.49 & 0.84 \\
    EP S-BERT + DA     & \textbf{0.84} & \textbf{0.90} & 0.90 & 0.84 & \textbf{0.78} & \textbf{0.57} & \textbf{0.85}\\
    \end{tabular}
    \caption{Mean F1 score for each class over a 5-fold cross validation as well as the weighted average (W/AVG) across all classes. The medical NER models (i.e., SciScpacy and BioBERT NER) only overlap with two classes in the HealthE label space. }
    
    % are only trained to output $\frac{2}{6}$ classes in the HealthE label space.
    \label{results}
\end{table*}

\subsection{Baselines and Related Works}
In this study we compare 6 baseline methods to our approach \textit{EP S-BERT}. Each of our entity classification baselines, namely GLoVe \cite{glove}, BERT \cite{bert}, Bio+Clinical BERT \cite{bioCbert} and S-BERT,  are used to produce embeddings for a given input entity which is then fed into a linear classification head. 

Additionally, to identify the capability of off-the-shelf Medical NER tools to extract entities of HealthE, we run two experiments using SciSpacy \cite{Neumann2019ScispaCyFA} and BioBERT NER \cite{biobert}.  We note this is not a one-to-one comparison with \textit{EP S-BERT}, which is fine-tuned to the HealthE label space, but rather an illustration of the limitations of existing Medical NER models and the datasets on which they are trained (i.e. clinical note and PubMed data).

% \textcolor{red}{Should reason why}
Given that S-BERT is not trained on any health data, we additionally evaluate \textit{EP S-BERT} with domain adaptation training using TSDAE  \cite{wang-2021-TSDAE}. We select TSDAE since it is a denoising autoencoder designed for use with the S-BERT model. This experiment, \textit{EP S-BERT + DA}, first enters a round of domain adaptive training on an additional ~40k unlabeled health texts scraped from the sources described in Section \ref{dataset_sec} before being fine-tuned to HealthE. This pre-disposes EP S-BERT to intermediate level health texts, which should better prepare it for understanding uncommon health terminology such as medication or treatment names.

\subsection{Results}
For each experiment, we report mean F1 score calculated over five-fold cross validation in Table \ref{results}. For each individual class, we report the F1 of the positive class only, as well as the weighted-F1 of the overall model. \textit{EP S-BERT} outperforms all baselines on the health entity classification task, highlighting the value of context pattern utilization. Additionally, we see a slight increase in performance from domain adaptation, specifically on the more medically-relevant topics such as DIS, MED and PHYS. This is to be expected as the standard S-BERT model is not trained on any health/medical specific texts. 

We also observe the Medical NER models struggling to classify DIS and MED entities in HealthE. We suspect that this is largely due to differing label spaces between HealthE and the medical corpora on which the Medical NER models were trained.  For example, in the advice sample ``You may also need this test if other tests, such as a blood glucose test, show you have \textit{low blood sugar}," HealthE has \textit{low blood sugar} labeled as DIS because it is a condition, whereas SciSpacy only extracts the chemical \textit{Glucose}. This label shift from existing Medical NER corpora marks an important distinction of HealthE as a general health advice corpus. 

% For example, SciSpacy is trained on the BC5CDNR corpus \cite{pmid27161011} which has 5818 DISEASE labeled samples from PubMed articles. HealthE has a Disease/Condition (DIS) label which is inclusive of conditions [cite the papers].

% \subsection{Comparison to BootstrapNet (SOTA)}
% \begin{itemize}
%     \item Explain how we tried to create a fair comparison to bootstrapnet by using different seed strategies, picking the best seed strategies
%     \item We also tested bootstrapnet with different embedding initializations (we can probably throw the results of this experiment to the appendix)
% \end{itemize}

% \begin{tabular}{lcccccc|c}
%   Model            & DIS  & MED  & FOOD & EXER & PHYS & OTH & W/AVG\\
% \hline
% GloVE              & 0.78 & 0.85 & 0.88 & 0.77 & 0.71 & 0.40 & 0.79 \\
% BERT               & 0.77 & 0.83 & 0.83 & 0.78 & 0.70 & 0.41 & 0.77 \\
% Bio+Clinical BERT  & 0.80 & 0.84 & 0.79 & 0.71 & 0.71 & 0.39 & 0.77 \\
% S-BERT             & 0.79 & 0.86 & 0.87 & 0.84 & 0.71 & 0.36 & 0.80 \\
% \hline 
% SciSpacy           & 0.59 & 0.44 & -    & -    & -    & -   &  0.51    \\
% BioBERT NER        & 0.00 & 0.00 & -    & -    & -    & -   &  0.00    \\
% \hline 
% EP S-BERT          & 0.81 & 0.89 & \textbf{0.92} & \textbf{0.88} & 0.75 & 0.49 & 0.84 \\
% EP S-BERT + DA     & \textbf{0.84} & \textbf{0.90} & 0.90 & 0.84 & \textbf{0.78} & \textbf{0.57} & \textbf{0.85}\\
% \end{tabular}

% \section{Results}
% \subfile{Sections/5results}

\section{Conclusion and Future Work}
%(There are two challenging subtasks within the HealthE dataset: 1) tools for existing health datasets do not effectively transfer to the HealthE task of medical advice labeling, 2) many entities in HealthE mean different things in different contexts, especially multi-token entities.)

Existing Medical NER tools perform poorly on HealthE, requiring new solutions to understand the unique HealthE label space. We show that transformer-based models perform well at entity classification and gain a significant performance boost from the encoding of entity context. The use of context/pattern embeddings is inspired by the low-resource task of Entity Set Expansion, a task under which HealthE could also be explored in future works. Future works may explore this problem in the context of medical NER by including a non-entity token label for all non-labeled tokens. Also, the performance increase from domain adaption highlights the potential for explicit integration of medical knowledge bases.

\bibliography{pmlr-sample.bib}

\newpage 
\appendix

\section{Data Annotation Details}\label{apd:first}

\subsection{Dataset}
In this work, we use a combination of two datasets, namely Online Health Advice (OHA) and Preclude \cite{preclude}. Both of these datasets contain textual health advice with entities (objects/phrases) annotated for the following six classes: (i) Food / Nutrient, (ii) Medicine / Supplement / Treatment, (iii) Disease / Condition, (iv) Exercise, (v) Vitals and Physiological Status, and (vi) Other. These classes are described in the annotation schema below. 

\subsubsection{Online Health Advice (OHA)}

\textbf{Collection: } The Online Health Advice (OHA) dataset consists of 5,600 samples of textual health advice collected from four online sources of information pertaining to health and general well-being: WebMD, Medline Plus, CDC, and Covid Protocols. 

\noindent \textbf{Annotation: }
These health advice statements were first annotated for usefulness. We define a health advice statement to be useful if it conveys some information, directly mentioned or implied, that is actionable. 2,536 out of the 5,600 statements were annotated as useful. Human annotators then manually extracted entities from the useful advice statements and labeled them as one or more of the six classes described in Section \ref{apd:annoDet} (Annotation Details). For the purpose of this work, these annotations were simplified so that each entity is annotated as one and only one of the six classes. 

\subsubsection{Preclude}
\noindent \textbf{Collection: } Preclude is an existing dataset consisting of 1,156 samples of textual health advice that primarily relate to food, exercise, lifestyle, and over-the-counter drugs. 790 of these advice statements came from 8 different mobile health apps and 366 of them came from WebMD, Yahoo! Health, MayoClinic, and Healthline.

\noindent \textbf{Annotation: } This dataset was originally annotated for medical conflict detection. 3 human annotators manually extracted entities from the advice statements and labeled them as positive, negative, or neutral. Then, for the purpose of this work, the manually extracted entities were then labeled by a human annotator as one the six classes described in Annotation Details. 

\subsection{Annotation Details}
\label{apd:annoDet}

A health advice statement in the corpus can have several entities, each annotated for different classes. However, for the purpose of clearly describing the classes, only entities in the class in question are labeled in the following examples. 

\subsubsection{Food / Nutrient (FOOD)}
Annotators were instructed to list objects or phrases that are food items or nutrients. Example advice statements from the dataset are listed below, in which labeled FOOD objects or phrases are in bold.
\begin{itemize}
    \item Finding \textbf{healthy food} choices on the road can be an adventure. Don't fill up on \textbf{low-fiber foods} at fast food chains, rest stops, or airports. Instead, pack a few \textbf{high-fiber snacks} for your trip to help keep you regular. Good choices include\textbf{whole grain crackers, dried or fresh fruit, fresh vegetables, or whole grain cereals}.
    \item A glass of \textbf{red wine} a day is good for you. The polyphenols in \textbf{green tea, red wine} and \textbf{olives} may also help protect you against breast cancer. The antioxidants may help protect you from environmental carcinogens such as passive tobacco smoke.
\end{itemize}

\subsubsection{Medicine / Supplement / Treatment (MED)}
Annotators were instructed to list objects or phrases that are a medicine, supplement, or treatment. Example advice statements from the dataset are listed below, in which labeled MED objects or phrases are in bold.

\begin{itemize}
    \item This \textbf{drug} should not be used with the following medications because very serious interactions may occur: \textbf{cisapride}, \textbf{dofetilide}.
    \item Certain medical conditions that increase bone breakdown, including kidney disease, Cushing's syndrome, and an overactive thyroid or parathyroid, can also lead to osteoporosis. \textbf{Glucocorticoids}, also known as \textbf{steroids}, also increase bone loss. \textbf{anti-seizure drugs} and long-term immobility because of paralysis or illness can also cause bone loss.
\end{itemize}

\subsubsection{Disease / Condition (DIS)}
Annotators were instructed to list objects or phrases that are the name of a disease or condition. Example advice statements from the dataset are listed below, in which labeled DIS objects or phrases are in bold.
\begin{itemize}
    \item Has your doctor said you have \textbf{high cholesterol}? Then you know you need to change your diet and lifestyle to lower cholesterol and your chance of getting \textbf{heart disease}. Even if you get a prescription for a cholesterol drug to help, you'll still need to change your diet and become more active for heart health. Start with these steps.
    \item See your health care provider if you think you have an \textbf{ACL injury}. Do not play sports or other activities until you have seen a provider and have been treated. 
\end{itemize}

\subsubsection{Exercise (EXER)}
Annotators were instructed to list objects or phrases that are names or types of exercise. Example advice statements from the dataset are listed below, in which labeled EXER objects or phrases are in bold.
\begin{itemize}
    \item See your health care provider if you think you have an ACL injury. Do not play \textbf{sports} or \textbf{other activities} until you have seen a provider and have been treated.
    \item  “Once you can do \textbf{stretching} and \textbf{strengthening exercises} without pain, you can gradually begin \textbf{running} or \textbf{cycling} again. Slowly build up distance and speed.
\end{itemize}

\subsubsection{Vitals / Physiological Status (VIT)}
Annotators were instructed to list objects or phrases that are the name of a vital or physiological status. This class also encompasses symptoms, behaviors, organs, and general state of health/being. Example advice statements from the dataset are listed below, in which labeled VIT objects or phrases are in bold.
\begin{itemize}
    \item If you or a loved one has signs of infection, talk to your doctor. Symptoms alone can’t tell whether klebsiella is the cause. So your doctor will test your spit, \textbf{blood, urine}, or other fluids to find out what type of bug is to blame.
    \item Ask your provider how often you should have your \textbf{a1c level} tested. Usually, testing every 3 or 6 months is recommended.
\end{itemize}

\subsubsection{Other (OTH)}
Annotators were instructed to list relevant objects or phrases that were not appropriate for the other five categories. This category exists in order to be exhaustive and inclusive of medically relevant objects or phrases that do not fit into the other five categories, but still may be important. Example advice statements from the dataset are listed below, in which labeled OTH objects or phrases are in bold. 
\begin{itemize}
    \item Do not \textbf{puncture} the canister or expose it to \textbf{high heat} or \textbf{open flame}. Keep all medications away from children and \textbf{pets}.
    \item Getting enough quality slumber may lower your pain and fatigue. Limit caffeine and alcohol and avoid tobacco. Eat your last meal of the day several hours before you go to sleep. Keep your bedroom comfortable and free of \textbf{electronics}.
\end{itemize}

\begin{table*}[]
\centering
\resizebox{\textwidth}{!}{\begin{tabular}{llllllll}
\multicolumn{1}{c}{ID} &
  \multicolumn{1}{c}{Text} &
  \multicolumn{1}{c}{FOOD} &
  \multicolumn{1}{c}{MED} &
  \multicolumn{1}{c}{DIS} &
  \multicolumn{1}{c}{EXER} &
  \multicolumn{1}{c}{VIT} &
  OTH \\ \hline
2592-2 &
  \begin{tabular}[c]{@{}l@{}}Some people with type 2 \\ diabetes can control their \\ blood sugar with healthy \\ food choices and physical \\ activity. But for others, a \\ diabetic meal plan and \\ physical activity are not \\ enough. They need to take \\ diabetes medicines.\end{tabular} &
  \begin{tabular}[c]{@{}l@{}}healthy \\ food choices, \\ diabetic meal \\ plan\end{tabular} &
  \begin{tabular}[c]{@{}l@{}}diabetes \\ medicines\end{tabular} &
  \begin{tabular}[c]{@{}l@{}}type 2 \\ diabetes\end{tabular} &
  \begin{tabular}[c]{@{}l@{}}physical \\ activity\end{tabular} &
   &
   \\ \hline
5515-6 &
  \begin{tabular}[c]{@{}l@{}}Knee, hip, and back \\ problems may put a \\ cramp in your walking \\ plans. Ask your doctor or \\ physical therapist for \\ advice before lacing up \\ your walking shoes. \\ Other problems that \\ might hinder walking \\ include balance issues, \\ muscle weakness, and \\ other physical disabilities.\end{tabular} &
   &
   &
  \begin{tabular}[c]{@{}l@{}}physical \\ disabilities\end{tabular} &
  walking &
  \begin{tabular}[c]{@{}l@{}}balance \\ issues, \\ muscle \\ weakness, \\ back \\ problems\end{tabular} &
   \\ \hline
5129-1 &
  \begin{tabular}[c]{@{}l@{}}Your doctor may use \\ x-rays to help confirm \\ the diagnosis and rule \\ out other types of arthritis. \\ X-rays show how much \\ joint damage has occurred.\end{tabular} &
   &
  x-rays &
  \begin{tabular}[c]{@{}l@{}}joint damage, \\ arthritis\end{tabular} &
   &
   &
  diagnosis \\ \hline
8053-4 &
  \begin{tabular}[c]{@{}l@{}}Burns can lead to many \\ complications, including \\ infection and bone and \\ joint problems. Because of \\ this, it’s a good idea to \\ always follow up with \\ your doctor.\end{tabular} &
   &
   &
  \begin{tabular}[c]{@{}l@{}}infection, \\ bone and \\ joint \\ problems\end{tabular} &
   &
  burns & \\ \hline
  8131-9 &
  \begin{tabular}[c]{@{}l@{}}Older adults may be more \\ sensitive to the side effects \\ of this drug, especially \\ dehydration and loss of \\ salts in the blood (such as \\ potassium, sodium).\end{tabular} &
   &
   &
  dehydration &
   &
  \begin{tabular}[c]{@{}l@{}}loss of salts\\ in the blood\end{tabular} &
  
\end{tabular}}
\caption{A sample of five health advice statements taken from HealthE with their associated labels. Each advice statement is accompanied by a unique identifier and each labeled health entity is listed in its respective category.}
\label{samples}
\end{table*}

A sample of five advice statements with their labels contained in HealthE is provided in Table \ref{samples}.
\section{Entity Topic Distribution}

\label{apd:second}
\begin{figure*}
    \centering
    \includegraphics[scale = 0.55]{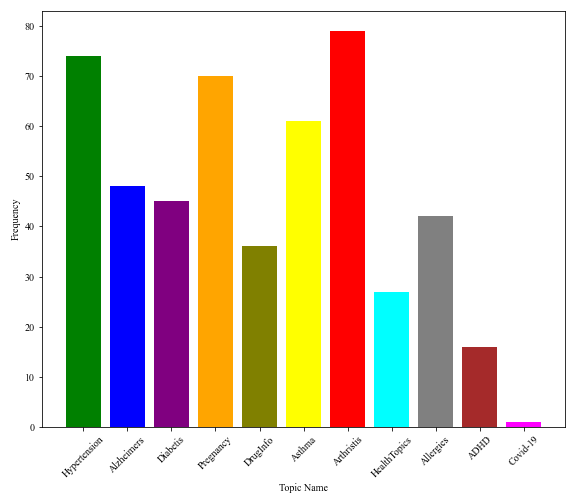}
    \caption{Entity Topic Distribution of Medical Advice Dataset}
    \label{Topics} 
\end{figure*}

To analyze the health topics covered by the full health advice dataset from which 
entities were extracted, we sample 500 advice statements and identify the article topic from which each statement was extracted. Figure \ref{Topics} shows the distribution of health advice topics discovered in this analysis. We find that most entities come from samples providing advice about Arthritis, Hypertension, and Pregnancy.

\end{document}